\pdfoutput=1

\documentclass[11pt]{article}

\usepackage[]{acl}

\usepackage{times}
\usepackage{latexsym}
\usepackage{booktabs}

\usepackage[T1]{fontenc}

\usepackage[utf8]{inputenc}

\usepackage[frozencache]{minted}
\usepackage{microtype}
\usepackage{subcaption}
\usepackage{graphicx}
\usepackage{tabularx}

\title{Dynatask: A Framework for Creating Dynamic AI Benchmark Tasks}

\author{Tristan Thrush$^\ddagger$\thanks{$\;\;$ TT and DK conducted most of the work for this paper when they were at Facebook AI Research.}, Kushal Tirumala$^\dagger$, Anmol Gupta$^\mathsection$, Max Bartolo$^\mathparagraph$, Pedro Rodriguez$^\dagger$, \AND Tariq Kane$^{\dagger \dagger}$, William Gaviria Rojas$^{\dagger \dagger}$, Peter Mattson$^\star$, Adina Williams$^\dagger$, Douwe Kiela$^\ddagger$\\\\
$^\ddagger$ Hugging Face;
$^\dagger$ Facebook AI Research;
$^\mathsection$ The University of Hong Kong;\\
$^\mathparagraph$ University College London;
$^\star$ Google;
$^{\dagger \dagger}$ Coactive AI;\\
{\tt\small dynabench@fb.com}
}

\begin{document}
\maketitle
\begin{abstract}
We introduce Dynatask: an open source system for setting up custom NLP tasks that aims to greatly lower the technical knowledge and effort required for hosting and evaluating state-of-the-art NLP models, as well as for conducting model in the loop data collection with crowdworkers. Dynatask is integrated with Dynabench, a research platform for rethinking benchmarking in AI that facilitates human and model in the loop data collection and evaluation. To create a task, users only need to write a short task configuration file from which the relevant web interfaces and model hosting infrastructure are automatically generated. The system is available at \url{https://dynabench.org/} and the full library can be found at \url{https://github.com/facebookresearch/dynabench}.
\end{abstract}

\section{Introduction}

Data is the backbone of NLP research. One of the most fruitful approaches for making progress on NLP tasks has historically been \textit{benchmarking}. Benchmarking is where the community adopts a high quality dataset for a particular task and tests various models against it to determine which is best. The process of benchmarking requires the effort of a large number of researchers, who collect and clean data, train and evaluate models, and work to understand model weaknesses. This process is iterative: once models perform very highly on the currently accepted community benchmark, another is created to push progress further. Taken as a whole, the benchmarking process is both notoriously difficult and expensive. This is due to a variety of facts: the community is a loose conglomeration of researchers with different areas of expertise, there is ever increasing need for larger datasets \citep{halevy-etal-2009-unreasonable}, and the AI community has historically under-valued~\citep{wagstaff-2012-machine} and under-invested in data collection and best practices \citep{dynabenchpaper, sambasivan-etal-2021-everyone, mattson2022dataperf}.

To make matters worse, in recent years, benchmarks have been saturating with increasing speed. Taking the trends from the greater AI community into account, it took MNIST~\citep{lecun1998mnist}, Switchboard~\citep{godfrey1992switchboard}, and ImageNet~\citep{deng2009imagenet} several years to saturate, and newer benchmarks such as SQuAD~\citep{rajpurkar-etal-2016-squad}, GLUE~\citep{wang2018glue}, and SuperGLUE~\citep{wang2019superglue} about a year. Because of this, data-centric approaches are gaining more attention \cite{data-centric-ai, mattson2022dataperf, lhoest-etal-2021-datasets,paullada2021data, luccioni2020data}. This trend is clear evidence of the urgency of finding a sustainable and data-centric way to support the full benchmarking ecosystem, from end-to-end, in a way that causes the least amount of friction for anyone who wants to use it.

In this paper, we introduce our answer to these issues: an easy-to-use, open source system that integrates the creation of benchmark datasets for any task, the selection of appropriate metrics, and the evaluation of models while natively supporting revisions to the benchmark as models saturate the original version. We share a unified library that enables these functionalities for the Dynabench platform \citep{dynabenchpaper}.

\begin{figure*}
\centering
\begin{subfigure}{.24\textwidth}
  \centering
  \begin{minted}[
    breaklines,
    gobble=4,
    frame=single,
    fontsize=\scriptsize
  ]{yaml}
    context:
    - name: context
      type: string
      placeholder: Enter context...
    input:
    - name: hypothesis
      type: string
      placeholder: Enter hypothesis...
    - name: label
      type: multiclass
      labels:
      - entailed
      - neutral
      - contradictory
      as_goal_message: true
    output:
    - name: label
    - name: probs
      type: probs
      reference_name: label
  \end{minted}
\end{subfigure}\hfill
\begin{subfigure}{.43\textwidth}
  \centering
  \includegraphics[width=\textwidth]{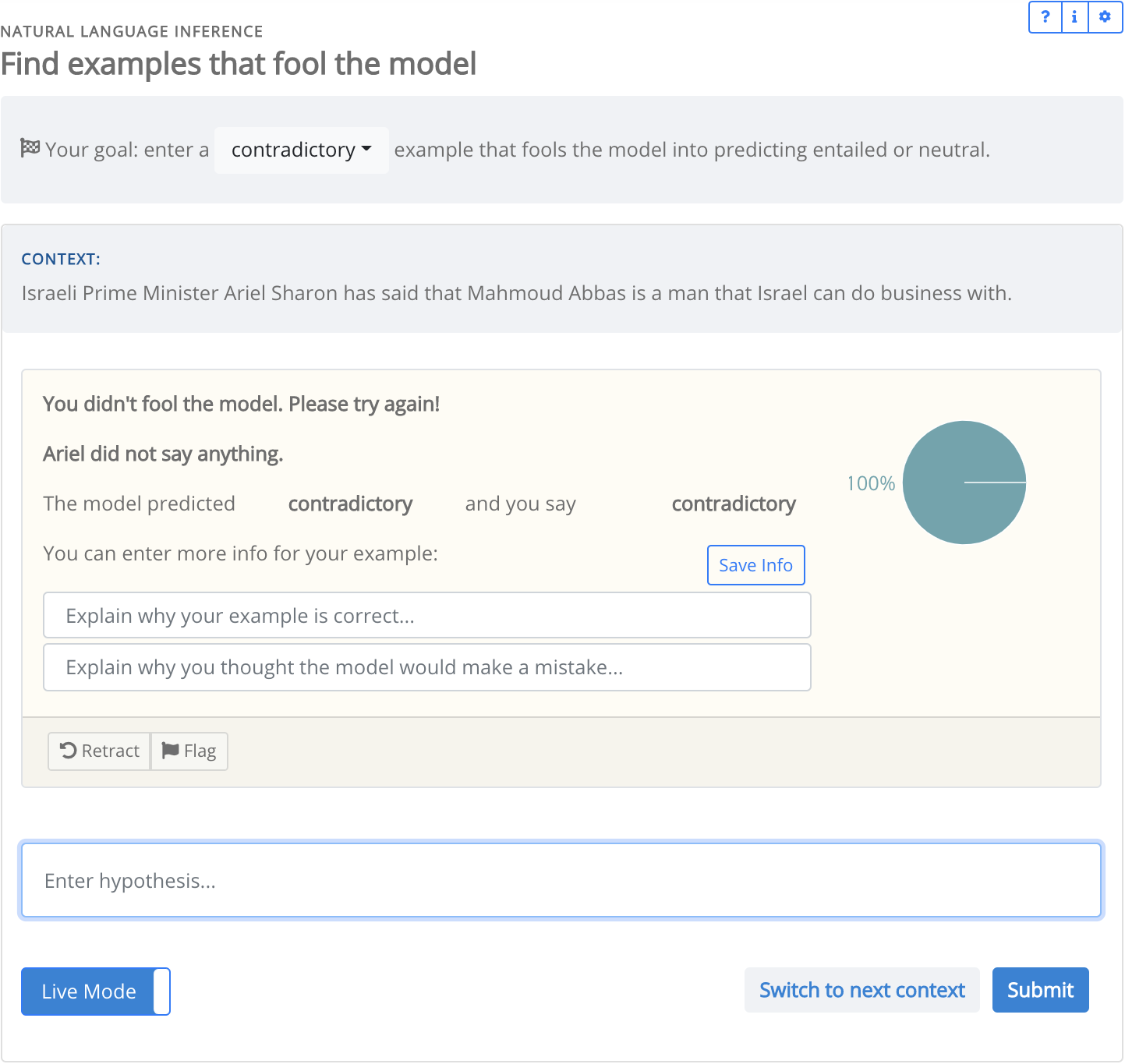}
\end{subfigure}
\begin{subfigure}{.32\textwidth}
  \centering
  \includegraphics[width=\textwidth]{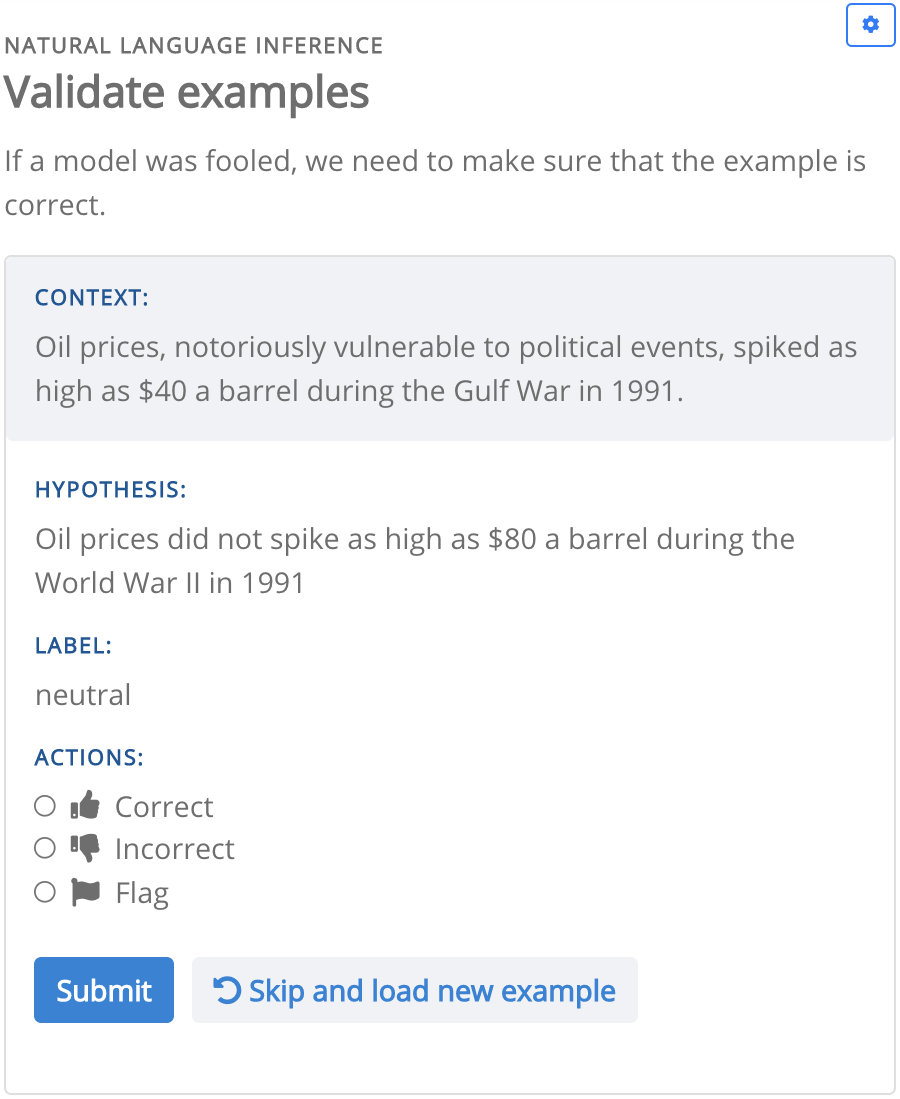}
\end{subfigure}
\par\bigskip
\rule[1ex]{\textwidth}{0.5pt}
\par\bigskip
\centering
\begin{subfigure}{.24\textwidth}
  \centering
  \begin{minted}[
    breaklines,
    gobble=4,
    frame=single,
    fontsize=\scriptsize
  ]{yaml}
    input:
    - name: image
      type: image
      display_name: image
    - name: labels
      type: multilabel
      labels:
      - Bird
      - Canoe
      - Croissant
      - Muffin
      - Pizza
    output:
    - name: labels
  \end{minted}
\end{subfigure}\hfill
\begin{subfigure}{.38\textwidth}
  \centering
  \includegraphics[width=\textwidth]{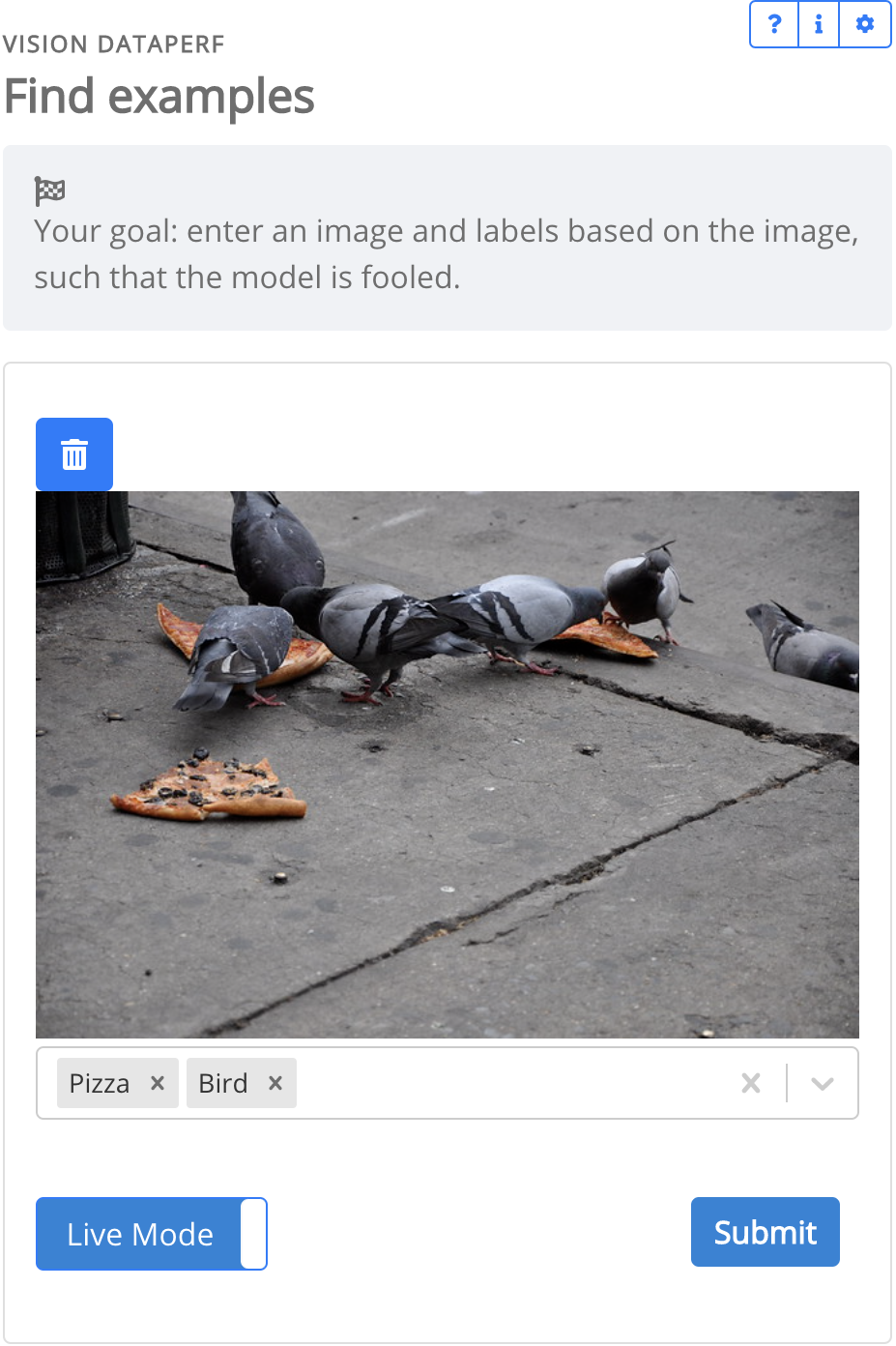}
\end{subfigure}
\begin{subfigure}{.37\textwidth}
\centering
\includegraphics[width=\textwidth]{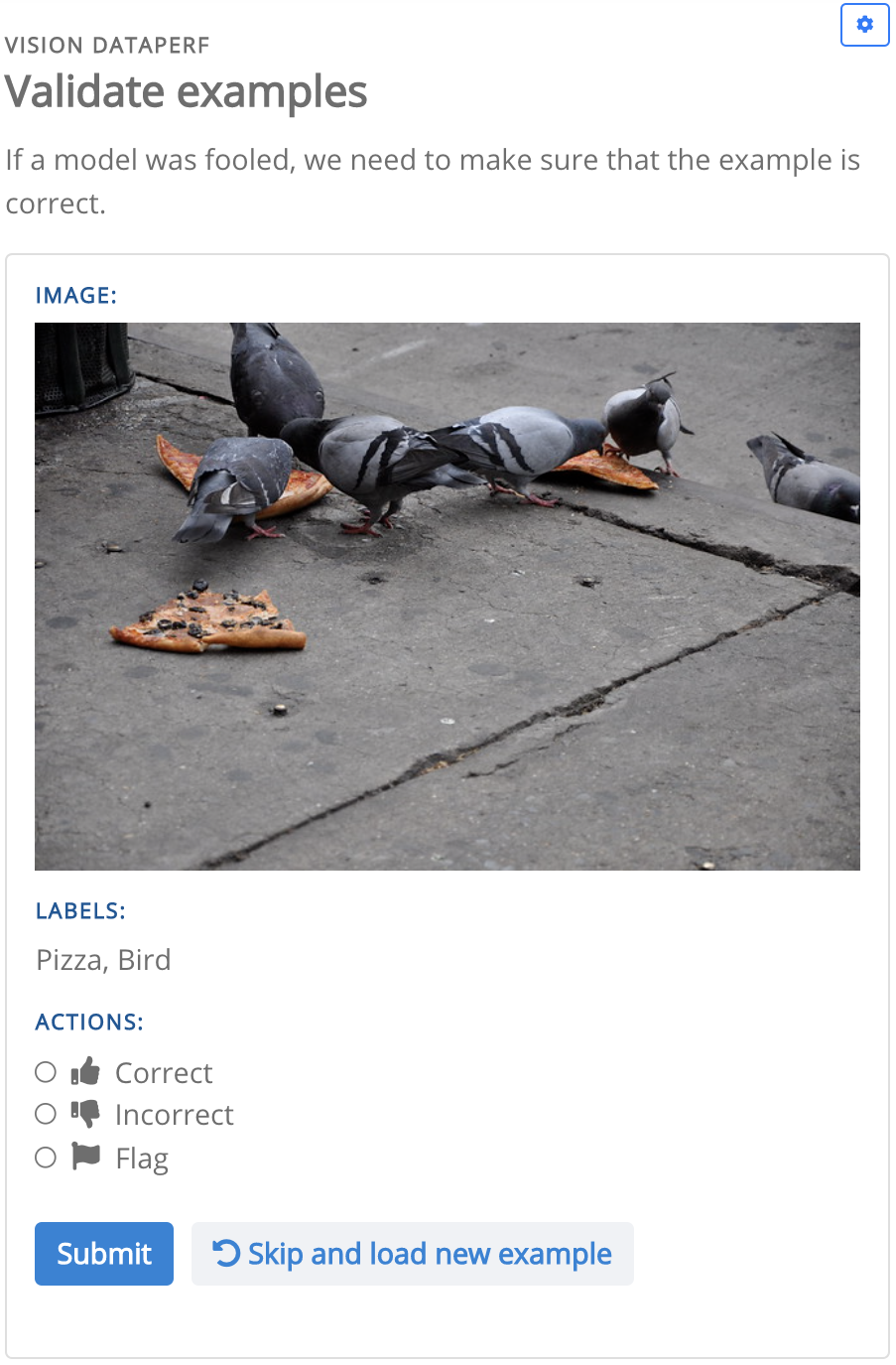}
\end{subfigure}
\caption{Two example config files and the data collection and validation interfaces they generate. Only config fields that impact the data collection interfaces are shown (e.g. metrics for model ranking are not shown). The Context and Input fields define the type of data that humans can enter. The Output field defines what models will output, given the Context and the Input. Crowdworkers are typically expected to provide the gold truth annotations for a task. In this case, Output will contain some of the object names from Input and Context. These gold truth annotations are removed from the Context and the Input before they are sent to models to get a model-in-the-loop output.\\\\ (\textbf{Top}) The config implements a natural language inference task. The first image is the collection interface, after a crowdworker submits their example and gets a model-in-the-loop response. The second image is the validation interface. For brevity, the metadata field in the config is omitted. This field is used to define the UI components for additional information, such as the ``Explain why your example is correct...'' input field.\\\\ (\textbf{Bottom}) The config implements an image labelling task. The first image is the collection interface, before a crowdworker submits their example. The second image is the validation interface with the same example.}
\label{fig:task_config_and_collection_interfaces}
\end{figure*}

\section{Background}

Dynabench was proposed as an open-source and community-driven platform to host dynamic benchmarks. The existing Dynabench tasks avoid saturation by leveraging crowdworkers who continually interact with state-of-the-art models. Crowdworkers either write examples that fool existing models~\cite{nie-etal-2020-adversarial}, or collaborate with generative models to increase example diversity \cite{bartolo2021models}. Each task is administered by one or more \textit{task owners} from the research community who collect data, make the competition's design decisions, select metrics, and configure the task's leaderboard.

\citet{dynabenchpaper} introduced Dynabench with four English language NLP tasks: Natural Language Inference~\cite{nie-etal-2020-adversarial}, Extractive QA~\cite{Bartolo2020BeatTA}, Sentiment Analysis~\citep{potts-etal-2020-dynasent} and Hate Speech Detection~\citep{Vidgen2020LearningFT}. In follow-up work,~\citet{ma2021dynaboard} updated Dynabench with additional leaderboard functionalities that allow task owners to upload task-specific models which are evaluated on each of the task's datasets, and can subsequently be included in model-ensembles that crowdworkers interact with. 
As the platform kept expanding, it became clear that Dynabench needed a scalable and configurable system for adding new tasks.

A \textit{task} is an essential concept in understanding our work. On Dynabench, a distinct task is a particular relationship between inputs and outputs.\footnote{Although, any user can set up a new task that is a duplicate of an existing one, with a duplicate config file.} Inputs and outputs are framed within some pre-specified format. For example, Natural Language Inference is a task on Dynabench. The input format is two strings and the output format is a classification label. The relationship between the inputs and outputs is defined by what humans would do when loosely instructed to treat the input strings as a context (sometimes called the ``premise'') and a hypothesis, and return a label for whether they think the hypothesis is entailed by the context. MNLI~\cite{williams2017broad}, SNLI~\citep{snliemnlp2015}, and ANLI~\citep{nie-etal-2020-adversarial} can be viewed as different datasets that instantiate the same task. \citet{schlangen-2021-targeting} takes a similar view.

\begin{figure*}
\centering
\begin{subfigure}{.29\textwidth}
  \centering
  \begin{minted}[
    breaklines,
    gobble=4,
    frame=single,
  ]{yaml}
    aggregation_metric:
      type: dynascore
    perf_metric:
      type: squad_f1
      reference_name: answer
    delta_metrics:
    - type: fairness
    - type: robustness
  \end{minted}
\end{subfigure}%
\begin{subfigure}{.71\textwidth}
  \centering
  \includegraphics[width=\textwidth]{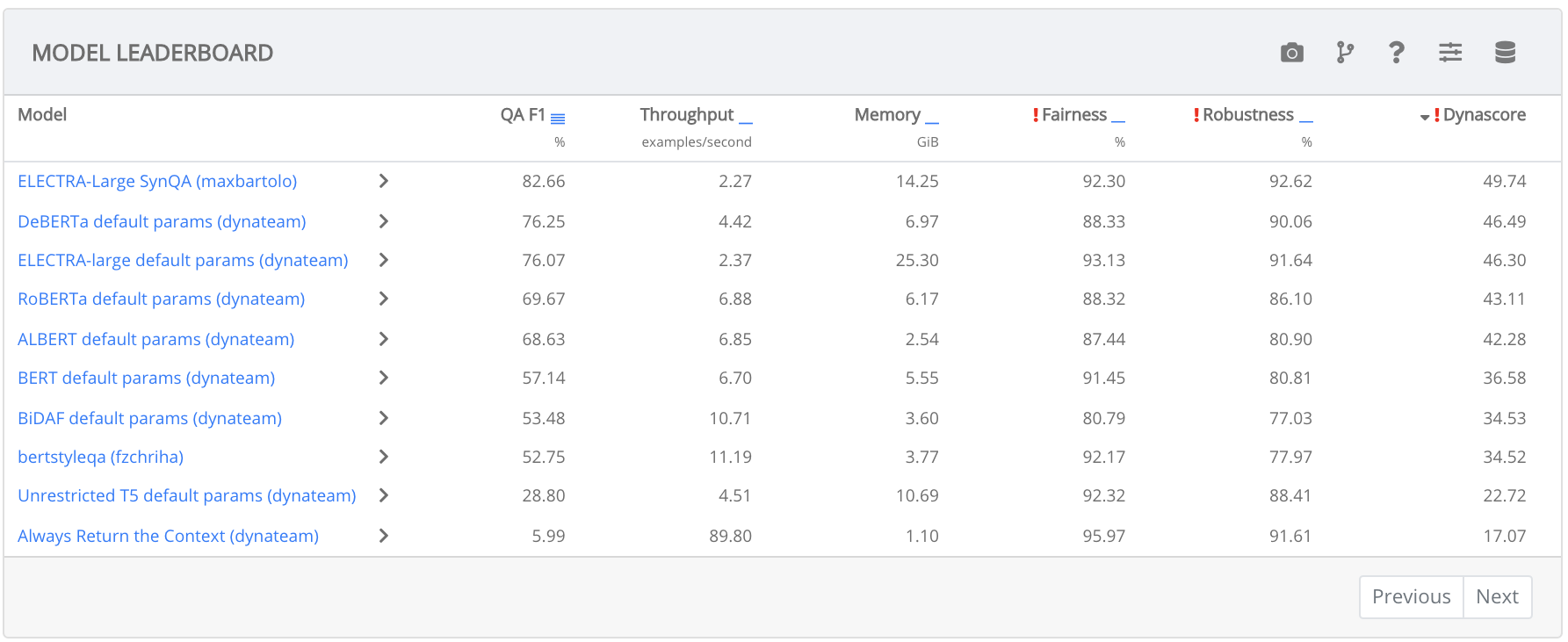}
\end{subfigure}
\caption{An example of a task config next to the generated model leaderboard. Only config fields that impact the leaderboard are shown. Throughput and memory do not need to be in the config; they are computed by default.}
\label{fig:task_config_and_leaderboard}
\end{figure*}

\section{Dynatask}

Before the introduction of Dynatask, adding a new task required close collaboration between task owners and the Dynabench team, and extensive software contributions to the Dynabench codebase. This paper presents a system that enables Dynabench to scale up to more tasks, including into multimodal and multilingual domains, without such requirements. Now, a task owner can create their own task page on Dynabench with a short task config file. The config file is used to automatically generate crowdworker data collection interfaces, as well as the model and dataset hosting/evaluating infrastructure. The data collection interfaces and hosting overlay existing services such as Amazon Mechanical Turk,\footnote{\url{https://www.mturk.com}} which provide a workforce and payment mechanisms, but do not provide crowdworker interfaces for dynamic model-in-the-loop data collection or their corresponding backends. In fact, local installations of Dynabench can be run on Mechanical Turk. Overall, a Dynabench task owner can set up and host:

\paragraph{Crowdworker data collection:} Task owners can configure interfaces for data collection. Models-in-the-loop can be optionally added, so crowdworkers can receive real-time model responses from their data (Figure~\ref{fig:task_config_and_collection_interfaces}).
\paragraph{Crowdworker data validation:} Task owners can configure interfaces for crowdworkers to label collected examples as correct or incorrect. (Figure~\ref{fig:task_config_and_collection_interfaces}).
\paragraph{Dynamic dataset metrics:} Metrics on the crowdworker data are computed, such as verified model error rate (vMER) \citep{nie-etal-2020-adversarial}. Crowdworker example leaderboards are displayed.
\paragraph{A train file leaderboard:} Task owners can enable users to upload training data files for the automatic creation, training, and evaluation of models in our evaluation cloud.
\paragraph{A dynamic and interactive model leaderboard~\citep{ma2021dynaboard}:} Task owners can configure a leaderboard, selecting from a variety of metrics to determine model performance. Owners can also upload new datasets, which triggers automatic evaluation for all of the user-uploaded models. Every leaderboard model can be interacted with in real-time. See Figure \ref{fig:task_config_and_leaderboard} for an example.
\paragraph{A model upload pipeline:} Once a new task goes live on Dynabench, our command line tool\footnote{\url{https://github.com/facebookresearch/dynalab}} allows anyone to create a handler script and upload models by following a few command line instructions. After models are uploaded, they are dockerized and deployed automatically. Models can be viewed on the leaderboard and put in-the-loop with crowdworkers for data collection.

\subsection{Task Configuration}

To become task owners, Dynabench users submit a short written proposal for their task which requires approval by an administrator. We are still developing procedures for how Dynabench accepts tasks; so far, we have reached out to have a discussion with the proposer before accepting their proposal and all non-spam proposals have been slated for acceptance. After approval, the task owner submits a task config file, which can be written in minutes. Once complete, the task is actively hosted on Dynabench; data collection, data validation, model hosting, and model evaluation starts immediately. A complete config file is the combination of a snippet in Figure \ref{fig:task_config_and_collection_interfaces} with that in Figure \ref{fig:task_config_and_leaderboard}.

The task config is a YAML file which allows someone to encode the specifications for their task---it can be viewed as a lightweight declarative programming language. Task owners can specify:

\textit{The datatypes of the task's inputs and outputs.} There are a variety to choose from, including String, String Selection, Multiclass, Multilabel, Probabilities, and Image. The datatype definition enables Dynatask to automatically construct the UIs for data collection, the dataset uploading and downloading infrastructure, and the model uploading and hosting infrastructure.

\textit{A variety of metrics} to understand the task's datasets and models. Several metrics can currently be computed for the leaderboard: Macro F1, F1 for Visual Question Answering, F1 for Question Answering, Accuracy, BLEU, robustness and fairness \citep{ma2021dynaboard}, memory usage, and example throughput. Task owners select or propose an aggregation metric, which combines results across multiple datasets and metrics to arrive at a ranking for the leaderboard. Currently, the only supported aggregation metric is the Dynascore \citep{ma2021dynaboard}, which combines metrics across datasets based on microeconomic utility \cite{ethayarajh-jurafsky-2020-utility} of user provided weights. Metrics can also be specified for model-in-the-loop data collection to judge whether a model's output matches that of a crowdworker (i.e., whether the model is ``correct''). Dynatask supports a variety of such metrics, including a string F1 threshold (for outputs that are strings), exact match, and simply asking the crowdworker whether the model was correct.

\textit{Other optional items}, such as messages and instructions that appear in crowdworker interfaces, and options for train-file leaderboards.

\begin{figure}
\includegraphics[width=\columnwidth]{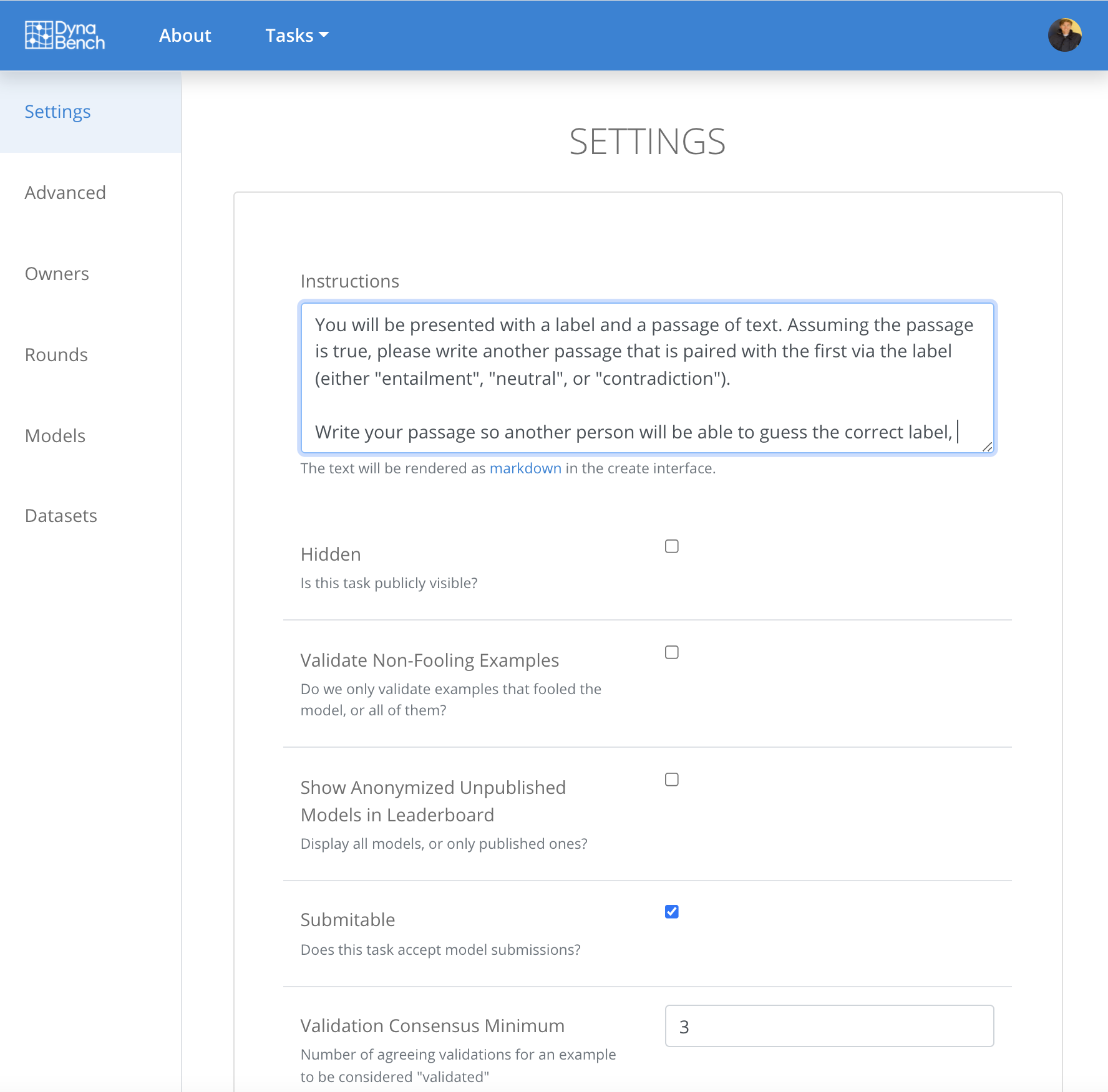}
\caption{The task owner interface for ANLI.}
\label{fig:task_owner_interface}
\end{figure}

\subsection{Options After Task Configuration}

\paragraph{Crowdworker Interfaces, Data Generation, and Data Evaluation:} Data collection interfaces are automatically hosted at \url{dynabench.org}. In order for Dynabench to scale, task owners source and pay crowdworkers themselves. If crowdworker management, compensation, and sourcing features are needed, an owner can clone Dynabench and run it on Mechanical Turk by hosting the data collection frontend using Mephisto.\footnote{\url{https://github.com/facebookresearch/Mephisto}} Task owners can upload context data for crowdworkers to use and download data collected from crowdworkers directly from the Dynabench web interface. Task owners can also initiate new rounds of data collection where they are free to upload entirely new contexts and models. As part of the data collection process, vMER, number of total collected examples, and number of validated examples are computed. Finally, task owners can alter instructions to crowdworkers at any time. They can also specify whether crowdworkers should validate non-model fooling examples, and provide a validation consensus threshold above which examples are considered fully validated. Figure \ref{fig:task_owner_interface} shows an example of the interface that task owners use to adjust settings.

\begin{figure}
\includegraphics[width=\columnwidth]{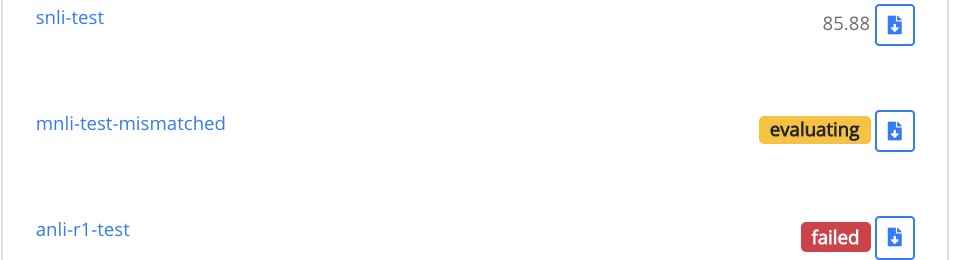}
\caption{A cross-section from a model card. The task owner has enabled downloading of model evaluation logs for each dataset via the buttons on the right.}
\label{fig:model_card}
\end{figure}

\paragraph{Model Submission, Interaction, and Evaluation:} Task owners can decide whether their task accepts model submissions, they can upload datasets for model evaluation, and they can download model evaluation logs for any dataset and model. Task owners can optionally allow users to download these logs to debug their models; see the example in Figure \ref{fig:model_card}.

\section{Decentralized Evaluation-As-A-Service}\label{sec:deaas}

Most task owners currently use the centralized Dynabench evaluation and model deployment server. With Dynatask, however, we offer a decentralized evaluation feature that will increase the platform's flexibility even further. With this feature, task owners can set up a Dynabench model deployment and evaluation server or select an existing one. To set up a new server, an owner only needs to follow our documentation, creating an AWS account and installing some Dynabench code along the way. Distributed hosting of model building and evaluation enables Dynatask to scale: no one organization needs to fund hosting for all of the models on Dynabench, and every owner of a model deployment and evaluation server can flexibly upload or take down models to suit their budget. It is also designed with re-usability in mind: several tasks can share the same evaluation servers. Task owners do not need to do any setup if they have permission to use an existing evaluation server.

\section{Case Studies of Tasks Enabled so Far}

Tables \ref{tab:dynabench_stats} and \ref{tab:task_types} provide an overview of Dynabench so far. In this section, we report on some use cases. Most of the following projects (besides Image Labelling and Open Domain QA) were added to Dynabench before the introduction of Dynatask, which took months of coding in every case. With Dynatask, they can all be implemented in minutes.

\begin{table}[t]
\centering
\begin{tabular}{lr}
\toprule
 Statistic      & Count \\\midrule
 Datasets Hosted & 191\\
 Unique Crowdworkers & 5,595\\
 Model Uploads & 589\\
 Data Collection Rounds & 38\\
 Tasks (incl. private) & 24 \\
 Examples Collected & 559,229\\
 Example Validations & 436,922\\
\bottomrule
\end{tabular}
\caption{Current Dynabench statistics.}
\label{tab:dynabench_stats}
\end{table}

\begin{table*}[t]
\small
\centering
\begin{tabularx}{\textwidth}{p{0.5\textwidth}  p{0.22\textwidth}  p{0.28\textwidth}}
\toprule
 Selected Dynabench Tasks      & Context and Input Types & Output Types \\\midrule
 Hate Speech Detection \tiny{\url{https://dynabench.org/tasks/hs}} & String, String, Multiclass & Multiclass, Probs\\
 Visual QA \tiny{\url{https://dynabench.org/tasks/vqa}} & Image, String & String\\
 Extractive QA \tiny{\url{https://dynabench.org/tasks/qa}} & String, String, String Select & String Select, Probs\\
 Open Domain QA \tiny{\url{https://dynabench.org/tasks/qb}} & String, String, String & String, Probs\\
 Natural Language Inference \tiny{\url{https://dynabench.org/tasks/nli}} & String, String, Multiclass & Multiclass, Probs\\
 Sentiment Analysis \tiny{\url{https://dynabench.org/tasks/sentiment}} & String, String, Multiclass & Multiclass, Probs\\
 Machine Translation \tiny{\url{https://dynabench.org/tasks/flores}} & String, String, String, String & String\\
 Image Labelling \tiny{\url{https://dynabench.org/tasks/vision-dataperf}} & Image, Multilabel & Multilabel\\
\bottomrule
\end{tabularx}
\caption{IO types from the task config, for some tasks on Dynabench. Tasks share the same building blocks.}
\label{tab:task_types}
\end{table*}

\textit{Hate Speech Detection:} There are a number of hate speech detection projects on Dynabench, where a model must label strings as hateful or not. Groups from Oxford, The Alan Turing Institute, The University of Sheffield, and Facebook AI own task pages that focus on collecting adversarial data~\citep{Vidgen2020LearningFT}, collecting emoji-based hate~\citep{kirk2021hatemoji}, and evaluating models on a large number of hate speech data perturbations. 

\textit{Visual Question Answering:} To combat saturating datasets for the VQA task, which is about answering a question based on an image, Facebook AI and Tecnológico de Monterrey introduced AdVQA~\citep{sheng2021human} using Dynabench. The task's model leaderboard has an additional adversarial VQA dataset from Microsoft and Tsinghua~\citep{li2021adversarial}.

\textit{Extractive Question Answering:} Groups from UCL and Facebook AI run SQuAD-style~\citep{rajpurkar-etal-2016-squad} extractive QA projects on Dynabench. The Adversarial QA~\citep{Bartolo2020BeatTA} project resulted in a popular dataset on the Hugging Face hub~\citep{lhoest-etal-2021-datasets}. Follow-up projects explored the generation of synthetic adversarial QA data~\citep{bartolo2021improving}, generative assistants in the loop to help annotators create examples~\citep{bartolo2021models}, and a study of how adversarial model-in-the-loop training data affects generalization out of domain~\citep{kaushik2021on}.

\textit{Open-Domain Question Answering:} A team at Facebook AI and The University of Maryland has started a model-in-the-loop data collection effort for the Quizbowl task \cite{rodriguez2019quizbowl,wallace-etal-2019-trick}, as well as a model leaderboard. The task is open domain question answering, where both the question and answer are strings.

\textit{Natural Language Inference:} The NLI dataset ANLI \cite{nie-etal-2020-adversarial} is currently a popular dataset on Hugging Face datasets~\cite{lhoest-etal-2021-datasets} and an ongoing Dynabench project. Groups from Facebook AI, UC Berkeley, and UNC have set up additional NLI projects on distinct Dynabench task pages. These projects have ranged from an analysis of the contents of adversarially collected development sets \citep{anlizing}, to an explication of the benefits of dynamic adversarial data collection over multiple rounds \citep{wallace2021analyzing}, to model and leaderboard hosting for a large number of robustness-perturbed NLI datasets. 

\textit{Sentiment Analysis:} In later rounds of their work, a team at Stanford used Dynabench to create a new adversarial sentiment analysis dataset, called Dynasent \citep{potts-etal-2020-dynasent}. They added prompts to their data collection interfaces to encourage crowdworkers to generate naturalistic and diverse data.

\textit{Large-Scale Machine Translation:} The Workshop on Machine Translation~\citep{wenzek2021findings} organizers created a Dynabench task page and hosted the FLORES benchmark competition~\citep{goyal2021the} of over 10,000 language pairs. It featured competitors from Microsoft, Huawei, Tencent, and Facebook, and individual competitors. The result of the competition was a BLEU increase of over 10 points on the full task. The owners used Dynabench for its leaderboard, model upload, and evaluation-as-a-service feature, without collecting data on the platform yet.

\textit{Image Labelling:} DataPerf~\cite{mattson2022dataperf} is a working group of the non-profit ML Commons, which focuses on dataset benchmarking for general AI. For their image labelling task hosted on Dynabench, they configured their task via the task config to accept training data file uploads. Users upload train files and models are automatically trained against them and evaluated in the evaluation cloud.

\section{Conclusion}

We introduced Dynatask, a collection of open source features in the Dynabench platform that empowers anyone to create and own a task on Dynabench with only a short config file. Dynabench started as an NLP project with only four English-only tasks. Since then, Dynatask has helped researchers produce several datasets and host competitions, expanding scalably into multimodal and multilingual domains with owners from various corners of the AI community.
Dynatask offers the functionalities of Dynabench to the broader research community by allowing them to easily create and host new AI tasks on the platform: it provides a one-stop shop for constructing datasets with or without models in the loop, hosting challenges and competitions, investigating the effects of models in the loop, characterizing distributional shift and continual learning, exploring annotator efficiency and expertise, and improving model robustness through collaboration with humans.

Finally, Dynabench is an open source, community-driven effort. Anyone who wants to add a new input/output type, a new metric, or any other new feature, need only submit a pull request. We hope that our our work can help enable new exciting scientific progress in data-centric AI research in general and dynamic (adversarial) data collection in particular.

\pagebreak

\bibliography{anthology,custom}
\bibliographystyle{acl_natbib}




\end{document}